\documentclass{styles/svproc}
\usepackage[dvipsnames]{xcolor}
\definecolor{DRNL}{RGB}{35, 206, 107}
\definecolor{DRL}{RGB}{254, 104, 71}
\definecolor{PLDagger}{RGB}{34, 108, 224}
\definecolor{PL}{RGB}{255, 165, 0}

\usepackage{graphicx}
\usepackage[inkscapeformat=png]{svg}

\usepackage{array} 

\usepackage{amsmath} 
\usepackage{subfigure}
\usepackage{url}

\begin{document}
\mainmatter              
\title{Strategic Sacrifice: Self-Organized Robot Swarm Localization for Inspection Productivity}
\titlerunning{Strategic Sacrifice}  
%
\author{Sneha Ramshanker\and Hungtang Ko \and
Radhika Nagpal}
\authorrunning{Ramshanker et al.} 
%
\tocauthor{Sneha Ramshanker, Hungtang Ko, and Radhika Nagpal}
\institute{Princeton University, Princeton NJ 08544, USA,\\
\email{s.ramshanker@princeton.edu, hk1581@princeton.edu, rn1627@princeton.edu}}

\maketitle              

\begin{abstract}
Robot swarms offer significant potential for inspecting diverse infrastructure, ranging from bridges to space stations. However, effective inspection requires accurate robot localization, which demands substantial computational resources and limits productivity. Inspired by biological systems, we introduce a novel cooperative localization mechanism that minimizes collective computation expenditure through self-organized sacrifice. Here, a few agents bear the computational burden of localization; through local interactions, they improve the inspection productivity of the swarm. Our approach adaptively maximizes inspection productivity for unconstrained trajectories in dynamic interaction and environmental settings. We demonstrate the optimality and robustness using mean-field analytical models, multi-agent simulations, and hardware experiments with metal climbing robots inspecting a 3D cylinder.

\end{abstract}
\section{Introduction}
\begin{figure}[tbp]
    \centering
    \includegraphics[width=0.8\textwidth]{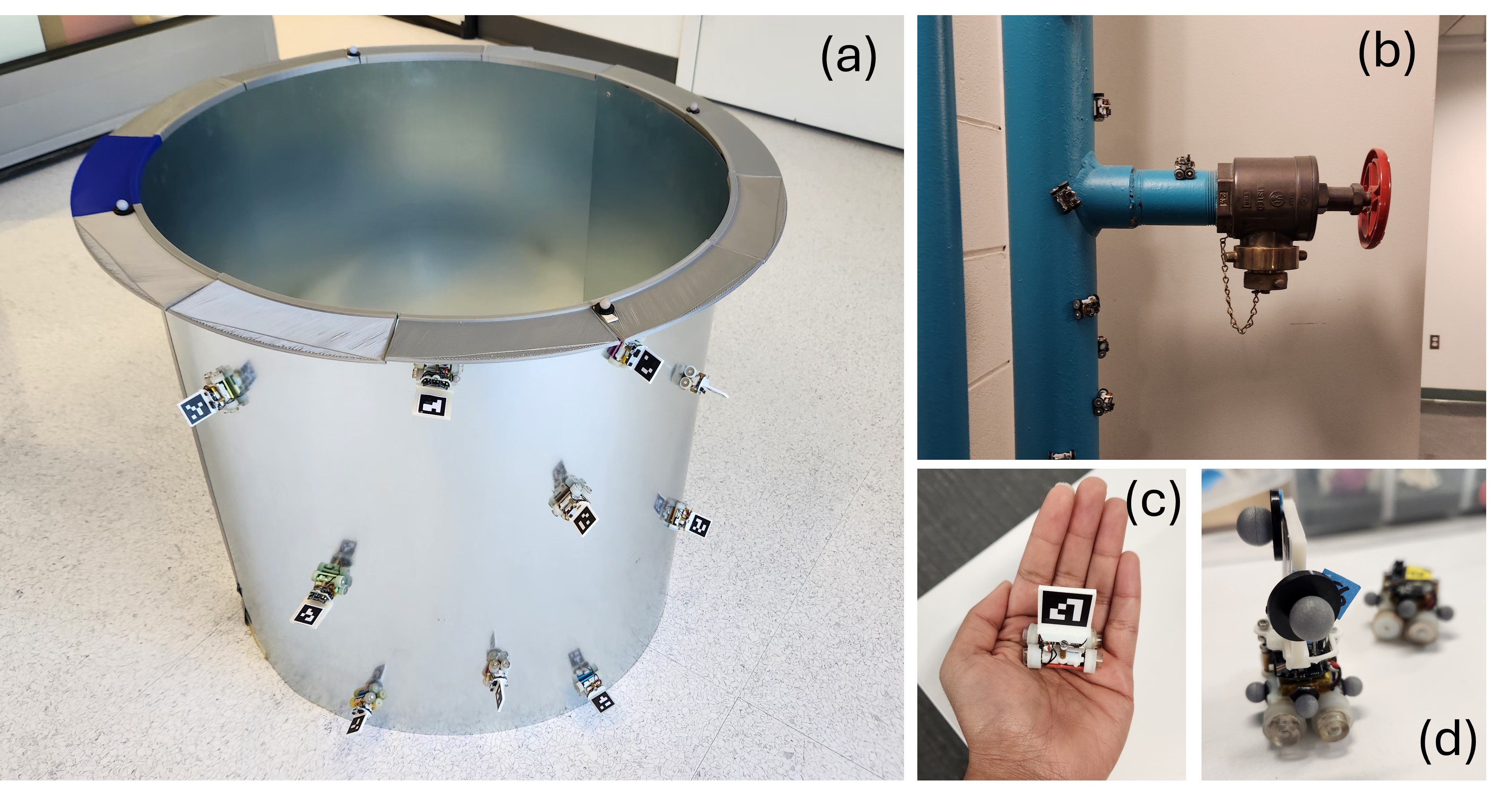} 
    \caption{Rovables \cite{dementyev2016rovables} (a) on a 3D metal cylinder, (b) on a piping system, and (c) in the palm of a human hand. (d) Rovables with markers for Vicon Motion Capture.}
    \label{fig:realrobots}
\end{figure}

Imagine a future where small robotic teams roam our infrastructure—bridges, pipelines, buildings, and satellites—detecting problems promptly, such as leaks and cracks (Figure \ref{fig:realrobots}(b)) \cite{halder2023robots,haghighat2022approach,correll2006collective,mclurkin2006speaking}. Teams of robots offer many advantages for inspection, including high parallelization, resilience to failure, and potentially low unit cost. However, for effective inspection, robots need to know where they are, a problem known as localization. In most indoor and remote locations, external localization mechanisms like GPS are unavailable or unreliable \cite{mautz2009overview}. Robots must autonomously perform localization, overcoming difficulties such as sensor noise, limited landmarks, and slippage. While the literature primarily addresses the localization problem from a single-robot perspective, with approaches such as SLAM achieving good accuracy \cite{cadena2016past}, these methods are computationally intensive. Computation is a finite and valuable resource; if most of an agent's computation is used for localization, little remains for productive inspection tasks like finding cracks, measuring vibration, and monitoring rust accumulation.

We explore if collaboration among robots can simplify this localization problem, allocating more computation for inspection. We draw inspiration from nature, where individuals sacrifice for the group's benefit. In anti-predator vigilance, some members watch for predators, allowing others to eat safely \cite{brugger2023looking,beauchamp2015animal}. Army ants form bridges out of their bodies, facilitating cargo transport across gaps and cracks \cite{malley2020eciton,mccreery2022hysteresis}. The vigilant animals forego eating, and the bridge ants do not carry cargo; instead, by sacrificing individual productivity, they enhance group success. What is especially interesting about these biological cases is that the fraction of sacrificers is self-organized and adapts to environmental demands.

We introduce a novel cooperative localization mechanism for robot swarms that leverages this idea of self-organized sacrifice for the group's benefit. While previous studies have used cooperation to enhance localization accuracy and enable localization in unknown environments \cite{prorok2012low,kurazume1994cooperative,nemsick2017cooperative}, these methods often increase computational requirements, constrain robot trajectories, and lack adaptability to changing environments. In our approach, individuals become dedicated localizers or inspectors, with this distribution self-organized based on local interactions. We demonstrate that this decentralized mechanism optimizes collective productivity in dynamic conditions, validated through theoretical models, numerical simulations, and hardware experiments. By deriving mean-field models (Sec \ref{sec: analytical model}), we prove that sacrificing agents as dedicated localizers improves swarm productivity.  We also show that the swarm can reconfigure based on local interactions to always maximize productivity. Using agent-based numerical simulations (Sec \ref{sec:Simulations}), we demonstrate that group productivity increases further through smarter collaboration. Hardware experiments (Sec \ref{sec:Hardware}) conducted with a swarm of 10 metal climbing robots inspecting a 3D metal cylinder (Figure \ref{fig:realrobots}) demonstrate the effectiveness of this approach in navigating complex physical environments. 
Moreover, these experiments show that the emergent behavior optimizes productivity amidst dynamically changing interactions. 

\section{Related Work}

Single-robot localization is extensively studied in robotics literature \cite{cadena2016past,thrun2002probabilistic}. The simplest approach is dead reckoning, where the robot uses internal sensor data like wheel encoders and IMU to estimate its position. While computationally inexpensive, dead reckoning suffers from error accumulation and requires regular external feedback for accuracy \cite{cadena2016past}. Probabilistic methods like particle filters and SLAM offer higher accuracy by incorporating external landmark and feature detection but demand heavy computational resources \cite{thrun2002probabilistic}. In this paper, we will not explore single-agent localization methods in detail. Instead, we will assume robots can achieve accurate localization at a high computational cost. An analysis of this accuracy-computation trade-off can be found in \cite{eyvazpour2023hardware}.

In multi-robot settings, various approaches improve localization using collaboration \cite{sullivan2018analysis}. A common strategy to boost accuracy is \textit{sharing personal estimates} and relative measurements \cite{prorok2012low,lajoie2021towards,luft2018recursive,klingner2019fault}. For instance, in \cite{prorok2012low}, each agent employs a particle filter (PF) combining personal dead reckoning with relative position estimates from neighboring agents. While these methods improve localization accuracy, they also come with an increase in computational costs, as all robots must process not only their individual measurements but also the estimates of their neighbors. Other collaborative localization (CL) approaches enhance accuracy by \textit{constraining swarm motion} \cite{kurazume1994cooperative,nemsick2017cooperative,pires2021cooperative,li2003distributed,tully2010leap}. For example, in \cite{mclurkin2006speaking}, robots disperse to form a landmark tree for navigation and in \cite{kurazume1994cooperative}, robots move in coordinated steps, with subgroups acting as stationary landmarks. However, the constrained motion can conflict with the motion required for inspection, increasing the complexity of inspection algorithms. By contrast, our approach is compatible with any unconstrained trajectory, offering the flexibility to execute any inspection algorithm. \textit{Heterogeneous CL} methods involve some robots performing more computation and sensing to localize others in the group \cite{nemsick2017cooperative,wanasinghe2015distributed,allotta2014cooperative,haldane2014detection}. For example, in \cite{nemsick2017cooperative}, observer robots with cameras localize beacon robots, sacrificing their computation for the group. However, in such heterogeneous approaches, the fraction of localizing agents is fixed apriori. Thus, the robot swarm cannot reconfigure itself if the environment or the needs of the group change. In our method, the fraction of sacrificing agents constantly adapts to maximize efficiency.

Our approach is also related to self-organized task allocation in biological and robot swarms \cite{pagliara2018regulation,albani2018dynamic,mayya2019closed}, building upon mean-field techniques commonly used in this area \cite{elamvazhuthi2019mean}. It maps to a task allocation problem where one task is computationally intensive localization, and the other is inspection. Unlike most studies, these tasks are tightly coupled in our approach, allowing the swarm to collectively achieve more than the sum of individual efforts via optimal task distribution.
\begin{figure}[tbp]
    \centering
    \includegraphics[width=\textwidth]{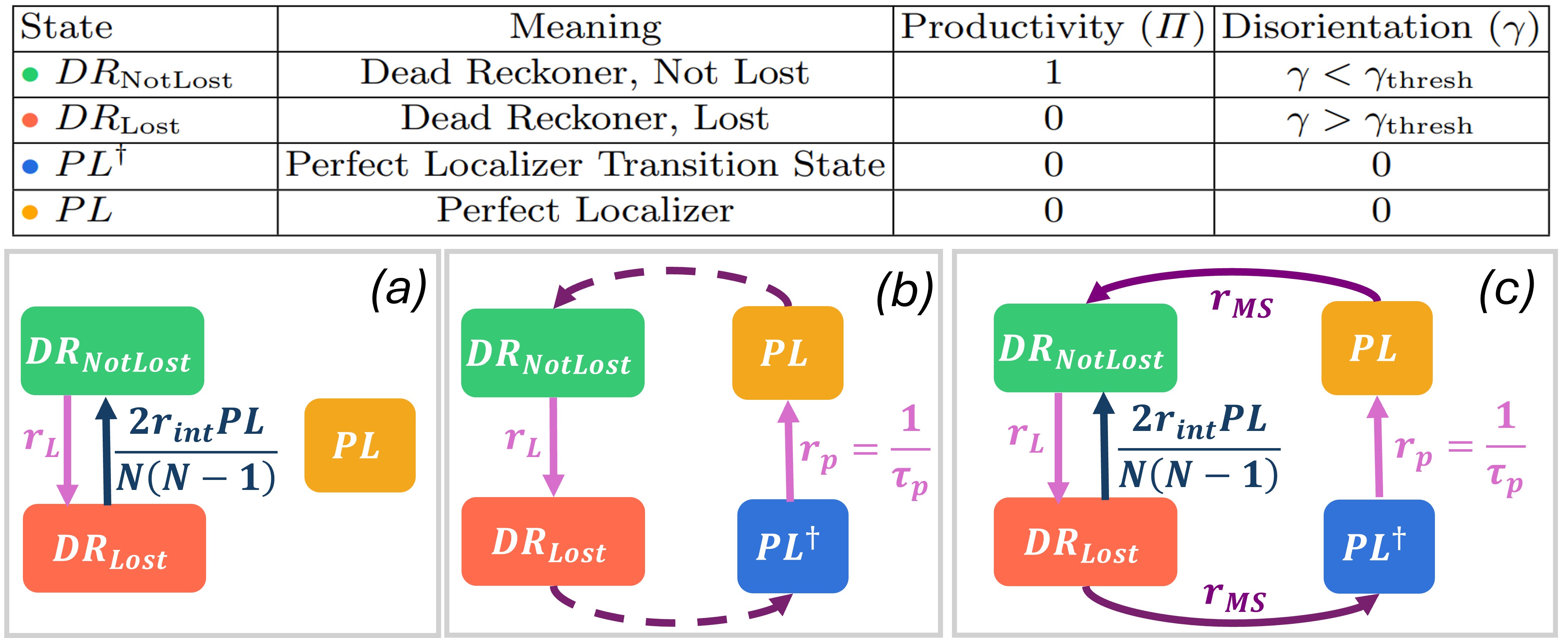}
    \caption{\textit{Top}: Summary of state properties. \textit{Bottom}: Transitions between $DR_{NotLost}, DR_{{Lost}}, PL^{\dagger}, PL$ for (a) fixed mode, (b) individual mode-switching, (c) collaborative mode-switching. Dashed arrows indicate instantaneous transitions while solid arrows indicate transitions at a finite rate.}
    \label{fig:RateEqDiagrams}
\end{figure}
\section{Problem Formulation}\label{sec:ProblemFormulation}
In our \textbf{collective localization} mechanism, agents operate in one of two modes:
\begin{itemize}
  \item \textbf{Dead Reckoner (DR) Mode:} Dead reckoners integrate IMU data to determine position and orientation. While computationally inexpensive, this approach suffers from error accumulation over time. Dead reckoners are productive when not lost but are increasingly prone to getting lost.
  \item \textbf{Perfect Localizer (PL) Mode:} Perfect localizers allocate all computational power to achieve perfect localization, sacrificing all inspection productivity. This assumption aligns with previous SLAM implementations on resource-constrained systems \cite{eyvazpour2023hardware}.
We introduce the $PL^{\dagger}$ state as a transitional phase, representing the startup cost for achieving perfect localization. In this state, localization is incomplete, and the agents cannot assist others. While undesirable, this is a necessary state for achieving perfect localization.
\end{itemize}
We represent the four states using the notation: $DR_{{NotLost}}$, $DR_{{Lost}}$, $PL^{\dagger}$, $PL$. A dead reckoner's lost status is characterized using a disorientation function, $\gamma= 1-\exp\left(-\frac{\left | \mathbf{\delta p} \right |}{\delta p_0}\right )$. Here, $\mathbf{\delta p}$ quantifies the localization error and $\delta p_0$ represents the characteristic localization error. Disorientation asymptotically approaches 1 as $\left| \mathbf{\delta p} \right |$ increases. We set a disorientation threshold, $\gamma_{{thresh}}$, to determine if an agent is lost: if $\gamma>\gamma_{{thresh}}$, the agent is considered lost. Designers can adjust the system's sensitivity to localization errors by changing $(\delta p_0, \gamma_{{thresh}})$. We assume agents can accurately estimate their disorientation, a strong assumption we aim to relax in the future. We use a step function to map dead reckoner disorientation to productivity, $\Pi$. If $\gamma<\gamma_{{thresh}}$, $\Pi = 1$; otherwise, $\Pi=0$. For perfect localizers, $\Pi=0$ and $\gamma=0$. The table in Figure \ref{fig:RateEqDiagrams} summarizes these states.

While perfect localizers are not directly productive, they play a crucial role in enhancing overall productivity by correcting the localization errors of dead reckoners through pairwise interactions. During such an interaction, the agents measure the relative position $\mathbf{p}_{\text{rel}}$ between them. The perfect localizer then shares its position, $\mathbf{p}_{\text{PL}}$, enabling the dead reckoner to calculate its true position as $\mathbf{p}_{\text{DR}} = \mathbf{p}_{\text{PL}} + \mathbf{p}_{\text{rel}}$. For now, we assume no uncertainty in $\mathbf{p}_{\text{rel}}$ or $\mathbf{p}_{\text{PL}}$, causing the dead reckoner’s localization error to collapse to zero after the interaction.

In the next section, we analyze these dynamics using analytical models, demonstrating the benefits of strategic sacrifice.

\section{Analytical Model using Mean Field \label{sec: analytical model}}
To model swarm dynamics and the effect of collaboration on productivity, we use mean-field ordinary differential equations (ODEs). Mean-field models simplify system dynamics by focusing on the evolution of mean values. This approach has been applied to various physical, social, and robotic systems \cite{elamvazhuthi2019mean,helbing2010quantitative}.

The swarm's occupancy configuration $n$ denotes the number of agents in each state: $n = \{ n_{DR}^{NotLost}, n_{DR}^{Lost}, n_{PL^{\dagger}}, n_{PL} \}$, where $n_{DR}^{NotLost}+n_{DR}^{Lost}+n_{PL^{\dagger}}+n_{PL}=N$ (total agents). We express the change in mean occupancy over time as $\frac{\mathrm{d} \left \langle n \right \rangle}{\mathrm{d} t}$ using ODEs. By solving for the steady-state configuration $\left \langle n \right \rangle_{ t\rightarrow \infty}$ where $\frac{\mathrm{d} \left \langle n \right \rangle_{ t\rightarrow \infty}}{\mathrm{d} t}=0$, we estimate mean productivity per agent using eq \ref{eq:generalProd}. The form of eq \ref{eq:generalProd} reflects that only non-lost dead reckoners are productive. We omit the $t\rightarrow\infty$ notation for clarity. $\frac{\Pi}{N}$ is normalized by experiment duration; e.g., if an agent is productive for half the experiment, $\frac{\Pi}{N} = 0.5$.

\begin{equation}
    \frac{\left \langle \Pi  \right \rangle}{N}= \frac{\left \langle n_{DR}^{NotLost}\right \rangle}{N}\label{eq:generalProd}
\end{equation}

We next find the steady-state dynamics for (1) Fixed Modes (2) Individual Mode-Switching, and (3) Collaborative Mode-Switching (Figure \ref{fig:RateEqDiagrams} a-c). 

\subsection{Fixed Modes, with Collaboration}\label{sec:AMfixedmodes}
We first assume a designer fixes the number of perfect localizers apriori ($N_{PL}$). The remaining agents are dead reckoners ($N_{DR}$). As shown in Figure \ref{fig:RateEqDiagrams}(a), dead reckoners transition to the lost state at a fixed rate $r_{L}$, which accounts for factors such as environmental slippage, observable landmarks, and sensor noise. Lost dead reckoners can return to the not-lost state via interactions with perfect localizers. At this stage, we assume a well-mixed system where every agent has an equal probability of interacting with any other agent in the swarm; the interaction rate between any two arbitrary agents is denoted as $r_{int}$. We want to determine the optimal number of perfect localizers that maximize productivity. The system's occupancy configuration is expressed by eq \ref{eq: FMIOde}. The right-hand side (RHS) terms are the rates of dead reckoners becoming lost and re-localizing via interactions, respectively. $n_{PL} = N_{PL}$, $n_{DR}^{{Not Lost}}+n_{DR}^{{Lost}}=N_{DR}$, and $N_{DR} + N_{PL} = N$. Solving eq \ref{eq: FMIOde} for the steady-state gives eq \ref{eq:appa_NMSI}. By computing the jacobian of eq \ref{eq: FMIOde}, we can show that the steady-state is a stable equilibrium.

\begin{align}
    \frac{\mathrm{d} \left\langle n_{DR}^{{Not Lost}}\right \rangle}{\mathrm{d} t} &= -r_{L}\left\langle n_{DR}^{{NotLost}}\right \rangle + \frac{2r_{int}}{N(N-1)}\left \langle n_{PL}\right \rangle \left \langle n_{DR}^{{Lost}}\right \rangle \label{eq: FMIOde} \\
    \frac{\left \langle \Pi \right \rangle}{N} &= \frac{1-\left( \frac{N_{PL}}{N}\right )}{1+\frac{r_L(N-1)}{2r_{int}}  \left( \frac{N}{N_{PL}}\right )} \label{eq:appa_NMSI}
\end{align}

 Differentiating eq \ref{eq:appa_NMSI} by the fraction of perfect localizers, $\frac{N_{PL}}{N}$, we find the $\frac{N_{PL}}{N}$ that maximizes the mean productivity per agent is eq \ref{eq:optimal_PL}.

\begin{equation}
    \left ( \frac{N_{PL}}{N} \right )^*=\frac{r_L(N-1)}{2r_{int}}\left ( -1+\sqrt{1+\frac{2r_{int}}{(N-1)r_L} }\right )
    \label{eq:optimal_PL}
\end{equation}

\begin{figure}[tbp]
    \centering
    \includegraphics[width = \textwidth]{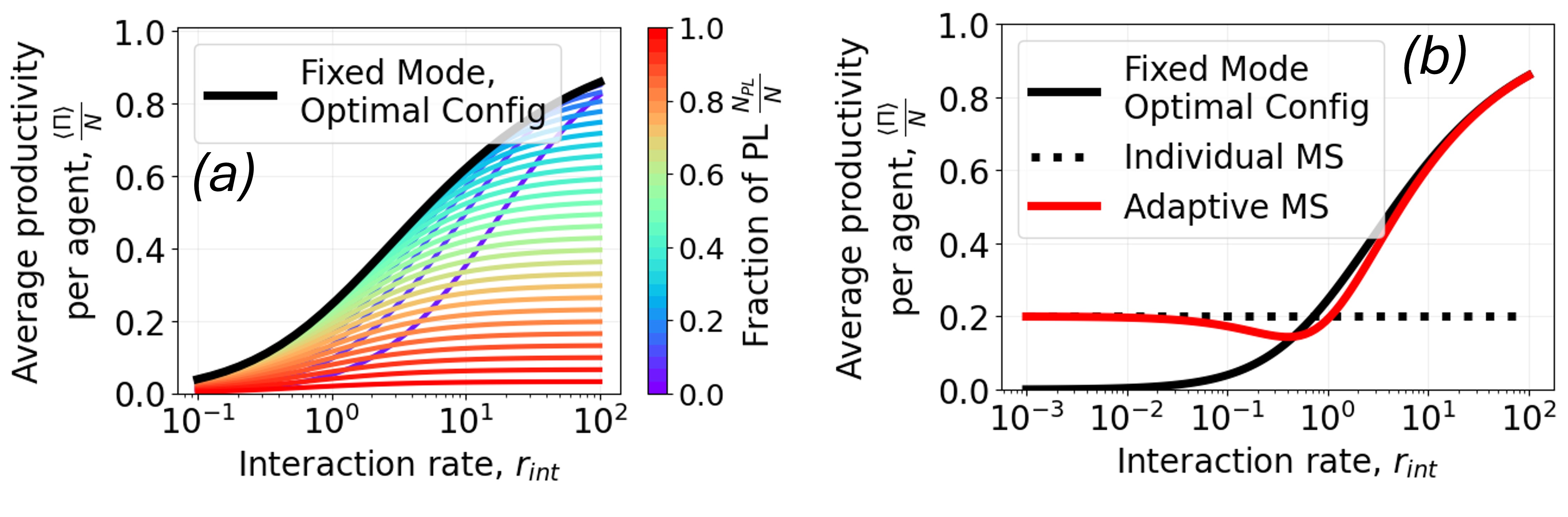}
    \caption{Mean-field predictions for (a) Fixed Mode (eq \ref{eq:appa_NMSI}), and (b) Mode-Switching. Solid black line is the fixed-mode productivity if initialized with optimal fraction of perfect localizers (eq. \ref{eq:optimal_PL}). Dotted line is the mode-switching productivity without collaboration (eq \ref{eq:appa_MSNI}). Red curve is the adaptive mode-switching strategy, where $r_{MS} = \frac{0.01}{r_{int}}$ (eq \ref{eq:appa_MSI}).}
    \label{fig:Analytical Model}
\end{figure}

Figure \ref{fig:Analytical Model}(a) shows $\frac{\left< \Pi\right>}{N}$ against the interaction rate, $r_{int}$, for different $N_{PL}$, where $N = 30$ agents and $r_{L} = 0.04$. Increasing $r_{int}$ enhances productivity per agent across all $\frac{N_{PL}}{N}$. Moreover, the optimal fraction of perfect localizers shifts with $r_{int}$: more perfect localizers are needed for maximum productivity at lower $r_{int}$ and fewer are required at higher $r_{int}$.

\subsection{Individual Mode-Switching, with No Collaboration}\label{sec:MSNITheory}

With fixed modes, average productivity per agent hinges on the initial mode configuration, i.e. the ratio between perfect localizers and dead reckoners (Figure \ref{fig:Analytical Model}(a)). Sub-optimal initial configurations can significantly reduce productivity. If the mode configurations are flexible, can swarms reconfigure to more optimal configurations by switching their modes?

We assume agents have the flexibility to switch modes independently but cannot interact with each other. Each agent optimizes productivity through periodic mode-switching: starting as a dead reckoner and transitioning to perfect localizer mode upon becoming lost. However, this transition incurs a cost as the agent must spend computation re-localizing in the $PL^{\dagger}$ state. This cost is captured by a fixed time-penalty $\tau_{p}$, which varies based on factors such as environmental stability and the number of observable features and landmarks. Once localized, the agent reverts to dead reckoning and resumes productivity until it gets lost again. Based on Figure \ref{fig:RateEqDiagrams}(b), the mean field is expressed by eq \ref{eq:MSNI_ODE}.

\begin{equation}
    \frac{\mathrm{d} \left \langle n_{DR}^{\text{Not Lost}} \right \rangle}{\mathrm{d} t}=r_p\left \langle n_{PL^{\dagger}} \right \rangle-r_L\left\langle n_{DR}^{\text{Not Lost}} \right \rangle
    \label{eq:MSNI_ODE}
\end{equation}

The RHS terms are the rates of mode-switching of perfect localizers to dead reckoners and vice versa, assuming instantaneous transitions. We approximate $r_p$ as $\frac{1}{\tau_p}$. Using the conservation of agent number, eq \ref{eq:appa_MSNI} gives the mean productivity per agent at stable equilibrium.

\begin{equation}
     \left< \frac{\Pi }{N} \right>=\frac{r
_p}{r_p+r_{L}}
\label{eq:appa_MSNI}
\end{equation}

Mode-switching allows lost agents to re-localize, regaining productivity for the collective. However, in this individualistic setup, productivity remains constant and independent of the interaction rate (Eq. 6 and Figure 3(b)). The system thus overlooks potential productivity gains from interactions, especially in scenarios where re-localization costs are high.

\subsection{Collaborative Mode-Switching} \label{sec:MSandCollab}

We combine mode-switching (Sec \ref{sec:MSNITheory}) with collaboration from our fixed modes scenario (Sec \ref{sec:AMfixedmodes}). This allows self-organization: in low interaction regimes, individuals mode-switch to be most productive, whereas, in high interaction regimes, fixed roles emerge. We examine whether this leads to optimal behavior. 

Agents operate according to Figure \ref{fig:RateEqDiagrams}(c), a superposition of Figure \ref{fig:RateEqDiagrams}(a) and \ref{fig:RateEqDiagrams}(b). We represent mode-switching using a rate, $r_{MS}$. The mean-field ODEs are 

\begin{subequations}\label{eq:MSI}
\begin{align}
    \frac{\mathrm{d} \left \langle n_{DR}^{{Lost}} \right \rangle}{\mathrm{d} t}&=r_L\left \langle n_{DR}^{{Not Lost}} \right \rangle-r_{MS}\left\langle n_{DR}^{{Lost}} \right \rangle-\frac{2r_{int}}{N(N-1)}\left \langle n_{PL} \right \rangle\left \langle n_{DR}^{{Lost}} \right \rangle \label{eq:example_a} \\
    \frac{\mathrm{d} \left \langle n_{PL^\dagger} \right \rangle}{\mathrm{d} t}&=r_{MS}\left\langle n_{DR}^{{Lost}} \right \rangle-r_p\left \langle n_{PL^{\dagger}} \right \rangle \\
    \frac{\mathrm{d} \left \langle n_{PL} \right \rangle}{\mathrm{d} t}&=r_{p}\left\langle n_{PL^\dagger} \right \rangle-r_{MS}\left \langle n_{PL} \right \rangle
\end{align}
\end{subequations}

In eq \ref{eq:MSI}(a), the first and last RHS terms mirror eq \ref{eq: FMIOde}. The middle term in eq \ref{eq:MSI}(a) and the first term in eq \ref{eq:MSI}(b) denote the rate of mode-switching of lost dead reckoners to $PL^{\dagger}$. The final term in eq \ref{eq:MSI}(b) and the first term in eq \ref{eq:MSI}(c) represent the rate at which $PL^{\dagger}$ transition to $PL$. In eq \ref{eq:MSI}(c), the last term is the rate at which perfect localizers mode-switch to dead reckoners. Using  conservation of the number of agents and solving for the steady state we get

\begin{equation}
    \frac{\left \langle \Pi \right \rangle}{N}=1-\left ( 2+\frac{r_{MS}}{r_p} \right )\frac{\left \langle n_{PL} \right \rangle_{t\rightarrow\infty}}{N} 
    \label{eq:appa_MSI}
\end{equation}

$\left< n_{PL}\right>_{t \rightarrow \infty}$ is the steady-state number of perfect localizers. Jacobian stability analysis and the Routh-Hurwitz Criteria confirm that this steady state is stable. If $r_{MS} >> r_{p}, r_{L}, r_{int}$, the system self-organizes to individual mode-switching, as seen by eq \ref{eq:appa_MSI} approaching eq \ref{eq:appa_MSNI}. Conversely, if $r_{MS} << r_{p}, r_{L}, r_{int}$, the system maintains fixed modes with an optimal number of perfect localizers, as evidenced by eq \ref{eq:appa_MSI} approaching eq \ref{eq:appa_NMSI} with $N_{PL} = N_{PL}^{*}$ (eq \ref{eq:optimal_PL}).

What is the optimal $r_{MS}$? In low $r_{int}$ scenarios, where perfect localizers provide minimal support, maintaining a high $r_{MS}$ prevents unnecessary loss in productivity that occurs by retaining dedicated perfect localizers. In high $r_{int}$ scenarios, where interactions alone can maintain the productivity of dead reckoners, a low $r_{MS}$ is preferable to avoid unnecessary mode-switching costs. Thus, if $r_{MS}$ is inversely correlated with $r_{int}$, the system can maximize productivity by adapting its mode-switching rate based on the agents' interaction rate. 

We call this form of collaboration \textbf{adaptive mode-switching}. We set $r_{MS}=\frac{\alpha}{r_{int}}$, where the proportionality constant, $\alpha$, determines the sensitivity of this dependence. Figure \ref{fig:Analytical Model}(b) illustrates the performance of adaptive mode-switching with $\alpha=0.01$, $\tau_p=100$, and $r_L=0.04$. It achieves near-optimal productivity for all interaction rates by converging to the optimal fixed mode configuration (Sec \ref{sec:AMfixedmodes}) when interaction rates are high and clipping to the individual mode-switching strategy (Sec \ref{sec:MSNITheory}) when interaction rates are low.

\section{Simulation: Multi-Agent Well-Mixed Scenarios \label{sec:Simulations}}

In this section, we conduct agent-based simulations to validate the accuracy of the mean-field analytical model (Sec \ref{sec: analytical model}) and explore the impact of initial conditions and non-steady state transients. Additionally, we introduce a \textit{smart collaboration} strategy to enhance productivity further.

We initialize $N=30$ agents, each transitioning to the lost state (${DR}_{Lost}$) after an interval of $\tau_L=3.46$ s in dead reckoning mode. Every $\tau_{int}$ seconds, two agents in the swarm randomly interact in a well-mixed manner. If a lost dead reckoner (${DR}_{Lost}$) interacts with a perfect localizer (PL), the dead reckoner corrects its localization error to 0. For each simulation run, we collect $\gamma$ and $\Pi$ for every agent at each timestep. The normalized average productivity per agent, $ \left< \frac{\Pi }{N}\right>$, is calculated as the cumulative productivity achieved by all agents, divided by $N$ and the total run duration, $\tau_{total}=200$ s. 

\begin{figure}[tbp]
    \centering
    \includegraphics[width=\textwidth]{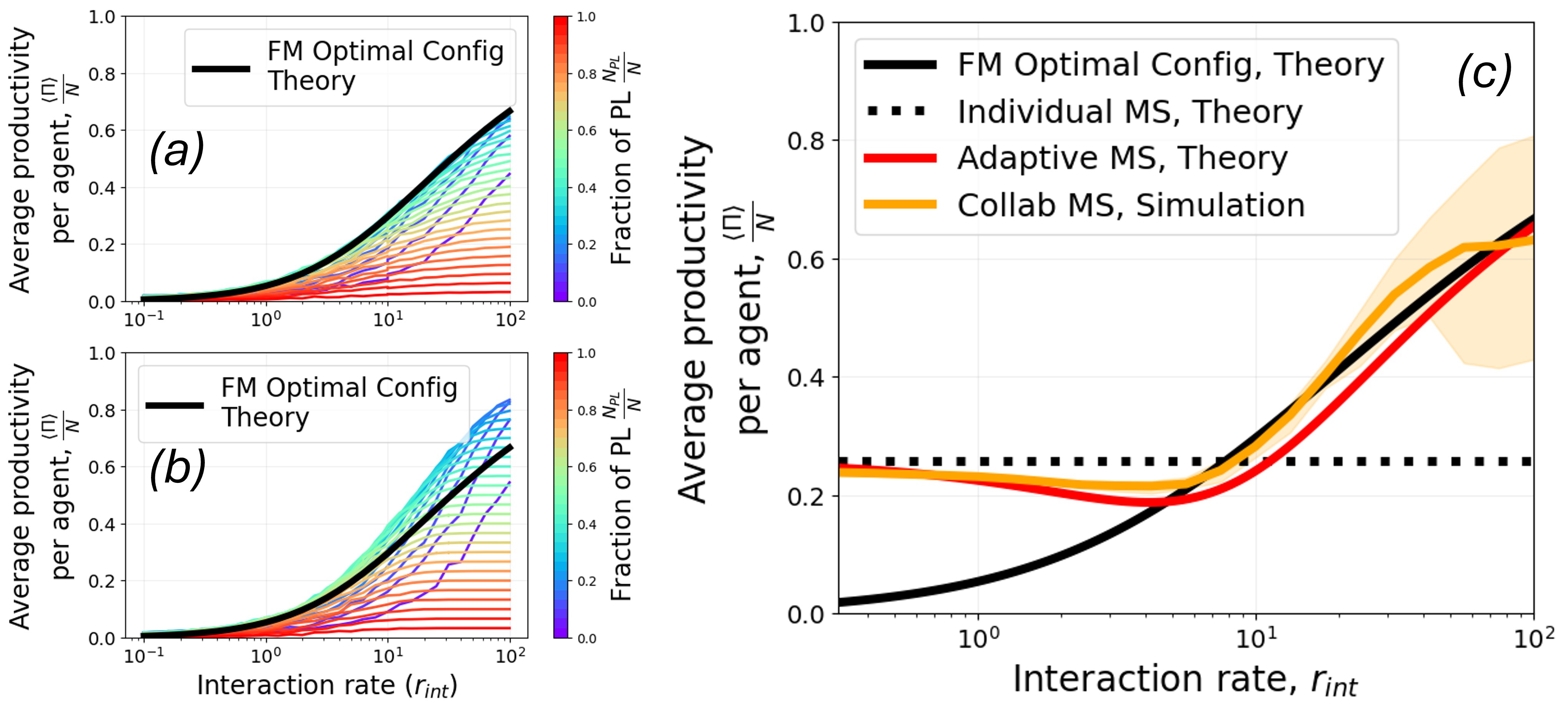}
    \caption{Agent-based Simulation Results. (a) Fixed Mode and Basic Collaboration, (b) Fixed Mode and Smart Collaboration, (c) Collaborative Mode-Switching}
    \label{fig:SimulationResults}
\end{figure}
We first consider the fixed modes scenario (Sec \ref{sec:AMfixedmodes}). Across 900 simulation runs, we vary the interaction rate $r_{int} = \frac{1}{\tau_{int}}$ and  fraction of perfect localizers $\frac{N_{PL}}{N}$. The results are shown in Figure \ref{fig:SimulationResults}(a), where each data point represents a single simulation run. We see close alignment between simulations and the analytical model, with minor variations due to transient behavior. 

Furthermore, smarter collaboration strategies significantly improve productivity. In Sec \ref{sec: analytical model}, only lost dead reckoners correct their localization estimates through interactions with perfect localizers. However, not-lost dead reckoners can also benefit from these interactions, correcting errors and maintaining productivity longer. Although challenging to model analytically due to non-linearity, numerical simulations confirm this effect. Figure \ref{fig:SimulationResults}(b) shows higher productivity in fixed mode configurations with this smart collaboration strategy, which we incorporate into subsequent simulations and experiments.

A second difference between the analytical model and multi-agent simulations occurs for collaborative mode-switching. In the analytical model (Sec \ref{sec:MSandCollab}), we demonstrate that the ideal mode-switching rate is inversely related to the interaction rate $r_{int}$. However, in the multi-agent simulations, agents lack knowledge of $r_{int}$ and must estimate it themselves. Each agent measures its local interaction rate, $\hat{r}_{int}$, using a simple counter and adjusts its mode-switching rate accordingly: $r_{MS}=\frac{1}{\hat{r}_{int}}$. If a dead reckoner mode-switches, it remains in the ${PL}^\dagger$ state for $\tau_{p}=10$ s before transitioning to a perfect localizer. Interactions between agents in the $PL^{\dagger}$ and dead reckoner states have no impact on the swarm. Like before, we conduct 900 simulation runs, exploring the interaction rate parameter space and initial mode configuration space. Unlike the fixed mode case, agents can change their modes if the initial configuration is sub-optimal. This setup allows us to observe the effect of initial conditions on performance variability.

Figure \ref{fig:SimulationResults}(c) presents the simulation results for collaborative mode-switching, which mostly outperforms the analytical adaptive mode-switching scheme with $\alpha=1$ across various $r_{int}$ values. The orange-shaded region represents the range of productivity across all initial mode configurations. Mode-switching reduces the sensitivity of average productivity to the initial mode configuration compared to fixed modes. However, at higher interaction rates, average productivity still exhibits some dependence on the initial configuration, although this effect is weaker. This is because swarms that start in optimal configurations for high interaction rates achieve high productivity from the outset. In contrast, swarms beginning in sub-optimal configurations must undergo transient mode-switching, temporarily reducing total productivity.

The well-mixed simulations above showcase the flexibility and effectiveness of collaborative mode-switching. However, real-world swarms have spatial dependencies and non-uniform localization errors not captured by these simulations. We address these complexities in the next section by testing with real robots.

\section{Hardware Experiments: 3D Inspection Swarm \label{sec:Hardware}}
\begin{figure}[tbp]
    \centering
    \includegraphics[width=\textwidth]{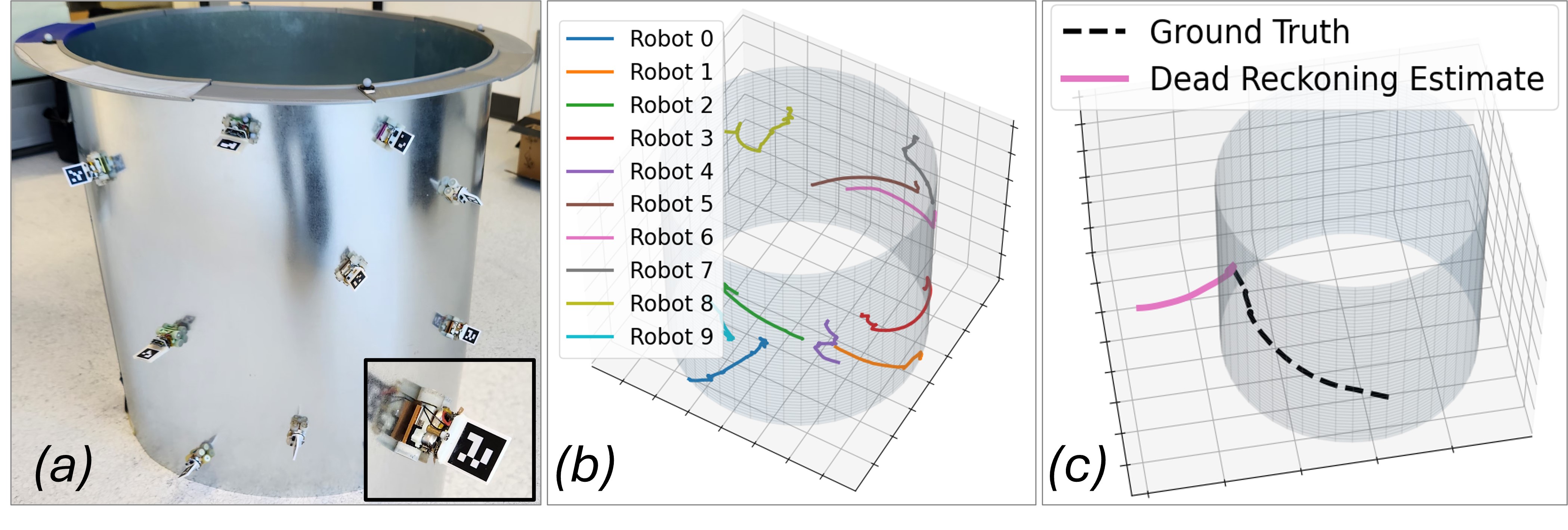}
    \caption{(a) 10 Rovables on a 3D cylinder and zoom in of a single Rovable, (b) Ground truth trajectories of the 10 Rovables, (c) Dead reckoning estimate of Robot 6.}
    \label{fig:hardwareExpConfig}
\end{figure}

Our group focuses on future applications of robot swarms for space structure inspection, aiming to reduce space debris and astronaut risks \cite{haghighat2022approach,chiu2023optimization}. We use a laboratory testbed of metal climbing Rovable robots \cite{dementyev2016rovables} inspecting a 3D cylinder structure (Figure \ref{fig:hardwareExpConfig}(a)). These robots are equipped with magnetized wheels for movement on metal surfaces and various onboard sensors, including radio communication and IMU for dead reckoning \cite{dementyev2016rovables}. Future Rovables will include cameras for detecting relative 6D pose using fiducial tags \cite{kalaitzakis2021fiducial}. Currently, we employ Vicon Motion Capture for line-of-sight vision over the curved 3D cylinder surface and to obtain ground-truth localization data.

We placed ten Rovables on a 3D metallic cylinder to simulate an inspection mission (depicted in Figure \ref{fig:hardwareExpConfig}(a)). This setup allows us to explore two significant factors. (1) We investigate the transition from well-mixed to spatial settings, where robots are more likely to interact with nearby neighbors, leading to dynamic interaction rates and interaction networks that are influenced by inspection trajectories. (2) We consider the variable rate at which robots become lost, influenced by physical factors like slippage, robot maneuvers, and sensor noise. 

To assess the impact of our collective localization method on swarm productivity, we the let robots perform random walks on the cylinder. IMU measurements were recorded, and ground truth positions were tracked using  Vicon (owing to Vicon limitations, trajectories were collected asynchronously). The trajectories of the ten robots are illustrated in Figure \ref{fig:hardwareExpConfig}(b). We tested various scenarios offline using this trajectory dataset, allowing us to compare the performance of different algorithmic strategies given the same experimental conditions.

Robots are initially classified as perfect localizers or dead reckoners, with perfect localizers utilizing ground truth position data from the Vicon system, and dead reckoners relying solely on onboard IMU data for localization. The disorientation observed in our experiments is influenced by physical factors such as slippage, sensor noise, and gravitational effects. Additionally, the rate at which robots become lost varies, with certain maneuvers and orientations leading to increased localization errors. Dead reckoning alone proves highly unreliable in the 3D curved setting, as evidenced by Figure \ref{fig:hardwareExpConfig}(c). We maintain $\gamma_{thresh}=0.4$ as before and set the localization penalty as $\tau_p = 20$ s; this parameter will eventually be time-varied and dependent on the perfect localizers' localization mechanism (e.g. particle filtering or SLAM). For now, we do not consider the productivity effects of different $\tau_p$. We also allow robots to know their localization error, so dead reckoners can detect when they are lost.

Interactions occur when robots establish line-of-sight vision of one another. During an interaction between a dead reckoner and a perfect localizer, the dead reckoner corrects its position using absolute position $\mathbf{p}_{\text{PL}}$ and relative position $\mathbf{p}_{\text{rel}}$ from the perfect localizer, reducing its localization error to zero (see Section \ref{sec:ProblemFormulation}). Currently, we do not account for errors in this process. We define the effective interaction rate as the frequency at which $DR_{NotLost}$ and $DR_{Lost}$ interact with $PL$ and $PL^{\dagger}$ robots. Robots determine this effective interaction rate by averaging their effective interaction count over a time window $\tau_{window}$. Averaging over $\tau_{window}$ allows for memory retention, enabling robots to make decisions based on recent observations while gradually forgetting older ones. The window size $\tau_{window}$ determines the sensitivity to interaction fluctuations. We set $\tau_{\text{window}} = 2\tau_p$. Mode-switching rate is inversely related to the average effective interaction rate. Unlike the well-mixed scenario where we could use the absolute interaction rate for mode-switching, in this context, we rely on the effective interaction rate to prevent the swarm from being trapped in local minima. 

\begin{figure}[tbp]
    \centering
    \includegraphics[width=0.95\textwidth]{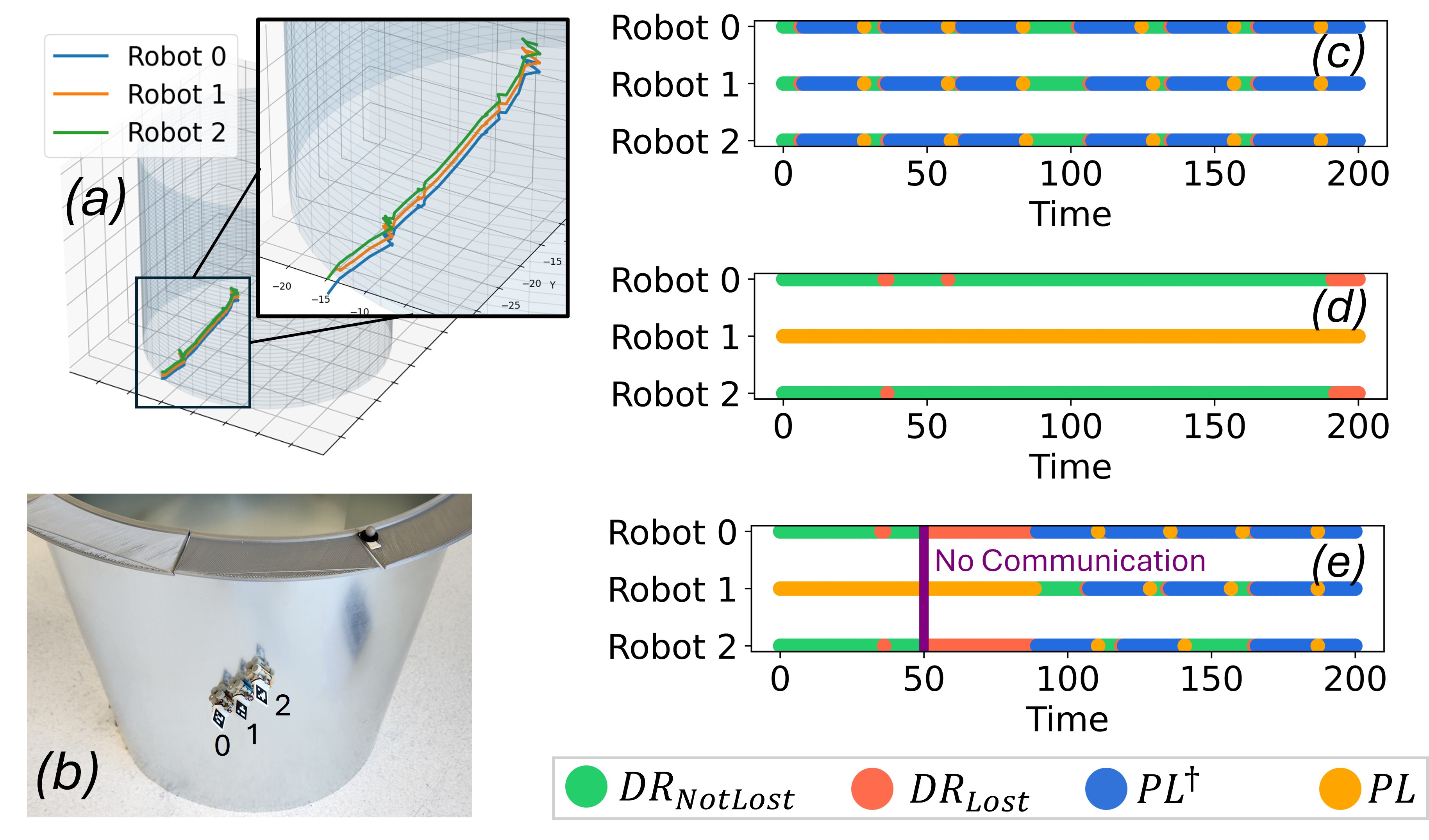}
    \caption{Collective localization of three Rovables in formation. (a) Ground truth trajectories of robots in formation, (b) 3 Rovables in formation on the physical cylinder. Robot mode time series for (c) individual mode-switching, (d) collaborative mode-switching, and (e) communication failure.}
    \label{fig:formation control}
\end{figure}

\subsection{Collective Localization of Three Robots in Formation}
We analyze $N = 3$ robots moving in formation to explore collective localization in a high interaction rate regime. However, formation control may not be the optimal strategy for inspection, as all robots cover the same subspace of the cylinder, limiting inspection coverage. Achieving formation control without vision is challenging, so we simulate it by shifting the Vicon data collected for a single robot (Robot 0) along the z-axis by 5 mm and 10 mm. The same IMU data is utilized for all three robots. To break symmetry, we vary the $\delta p_0$ parameter: [1.0, 1.3, 1.5]. In the spatial setting, the choice of formation also influences productivity. For the formation used in this case study, the middle robot interacts with two edge robots, but the edge robots do not see each other due to occlusion. Figure \ref{fig:formation control}(a) and (b) display the trajectories and provide a visual representation of formation control on a physical cylinder. We examine three distinct cases:

\begin{itemize}
    \item \textbf{Individual Mode-Switching}\label{sec:MS_NI_FC_test} This scenario models individual robot localization without any collaboration. As depicted in Figure \ref{fig:formation control}(c), the blue-yellow-green snake-like pattern demonstrates the robots alternating between productive phases and incurring time penalties in the $PL^{\dagger}$ state. This behavior aligns with the periodic mode-switching pattern discussed in section \ref{sec:MSNITheory}, resulting in an average productivity per agent of 0.22.

    \item \textbf{Collaborative Mode-Switching}: Robots collaborate and are initialized as [DR, PL, DR]. Robot 1 interacts with Robot 0 and Robot 2 throughout the simulation. As illustrated in Figure \ref{fig:formation control}(d), the swarm organizes to designate Robot 1 as the perfect localizer, while Robots 0 and 2 remain dedicated dead reckoners. Robots 0 and 2 use Robot 1 like a GPS, continuously obtaining position information through interactions. The total productivity per agent reaches 0.63, nearly three times higher than without collaboration.

    \item \textbf{Communication Breaks} We show the system's robustness to interaction rate changes by implementing a scenario where robots can interact for the first 50 seconds, after which communication is disabled. As shown in Figure \ref{fig:formation control}(e), initially, the swarm designates Robot 1 as the dedicated perfect localizer. However, after communication is disrupted, it dynamically self-organizes into a mode-switching behavior. Note, this behavior occurs purely via local interactions. Between 50 and 80 seconds, despite being unproductive, the robots persist in dead reckoning and perfect localizer mode. This hysteresis behavior is influenced by the memory window $\tau_{window}$. The effect of the hysteresis is more profound for larger $\tau_{window}$, but the swarm is also more resistant to fluctuation in interaction rate. Adjusting the memory window size allows system designers to modify swarm behavior. 
\end{itemize}

This simple case study highlights several intriguing properties. First, in scenarios where robots do not collaborate, we illustrate how mode-switching enables robots to sustain productivity. Second, within the high interaction rate regime, the swarm realizes a substantial productivity increase by allocating one robot to support the maximum productivity of the remaining two robots. Lastly, the swarm demonstrates resilience to failures, such as signal jamming, reconfiguring to mode-switching when the interaction rate declines.

\subsection{Coverage by Ten Random-Walking Robots}
We explore a complex scenario where all 10 robots follow unconstrained trajectories, randomly traversing the cylinder for coverage. We generate 30 runs by randomly rotating and translating the 10 original trajectories (Figure \ref{fig:hardwareExpConfig}(b)) around the cylinder's axis of symmetry. This approach produces complex interaction profiles, leading to dynamic interaction networks and varying interaction rates within and across experimental runs. We then analyze the productivity under Fixed Modes, Individual Mode-Switching, and Collaborative Mode-Switching.

All robots are initialized as dead reckoners, with $\delta p_0=1$ and $\tau_p = 20$ s. For Fixed Mode and Individual Mode-Switching, results are consistent across all runs as these modes are unaffected by interaction profiles. Under Fixed Mode, the productivity per agent is low at $0.02$, as robots quickly become lost. Individual Mode-Switching yields slightly improved productivity, reaching $0.138$ per agent. However, Collaborative Mode-Switching, which is sensitive to interaction profiles, performs the best for all 30 runs, achieving an average productivity per agent of $0.278 \pm 0.046$.

\begin{figure}[tbp]
    \centering
    \includegraphics[width = 0.85\textwidth]{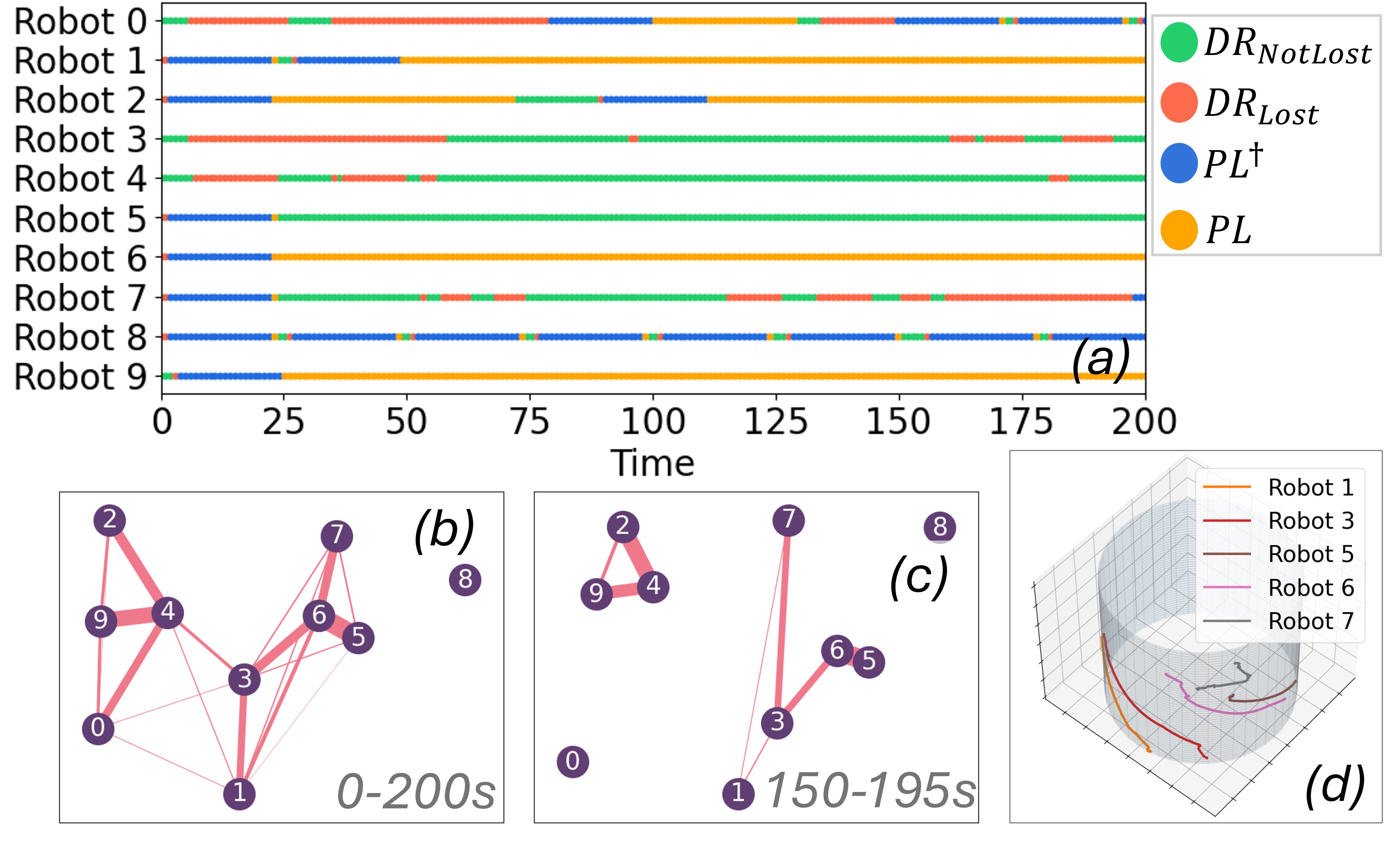}
    \caption{10 Random-Walkers Results. (a) Robot mode time series. Interaction network (b) from 0-200 s (full experiment duration) and (c) 150-195s.  Edges represent interactions between robots within the specified time frames, with edge weight proportional to the number of interactions. (d) Ground truth trajectory of Robots 1, 3, 5, 6, 7.}
    \label{fig:10TrajResults}
\end{figure}

To gain insight into the underlying dynamics and understand how interactions improve productivity, we examine the Robot Mode Time Series and Interaction Networks (Figure \ref{fig:10TrajResults}) of a single run. The productivity per agent for this run under collaborative mode-switching is 0.304.

\begin{itemize}
\item \textbf{Robot 8 is isolated from the rest of the swarm} throughout the run, as seen in Figure \ref{fig:10TrajResults}(b). Operating on the opposite side of the cylinder, it lacks line-of-sight connectivity with other robots. Consequently, as depicted in Figure \ref{fig:10TrajResults}(a), Robot 8 exhibits mode-switching behavior similar to that shown in Figure \ref{fig:formation control}(c). Robot 0 encounters similar isolation around $t = 150$ s, as illustrated in Figure \ref{fig:10TrajResults}(c). In Figure \ref{fig:10TrajResults}(a), we observe Robot 0 transitioning to pure mode-switching following a period of hysteresis.

\item \textbf{Robot 6 interacts with Robots 7, 5, 3, and 1}. From Figure \ref{fig:10TrajResults}(a), the swarm self-organizes to sacrifice Robot 6 as the perfect localizer, allowing Robots 3, 5, and 7 to reap the productivity benefits (similar to Figure \ref{fig:formation control}(d)). The behavior of Robot 1 is complex as it interacts with several other robots.

\item \textbf{Robot 4 exhibits an interesting individual behavior}, consistently remaining a dead reckoner while benefiting from other perfect localizers (Figure \ref{fig:10TrajResults}(a)). This strategy, while reasonable on the individual level, is sub-optimal for the group. Given Robot 4's central location (Figure \ref{fig:hardwareExpConfig}(b) and \ref{fig:10TrajResults}(b)), it could assist more robots as a perfect localizer. If we rerun the trial with Robot 4 starting as a perfect localizer, productivity per agent increases to 0.38, with Robot 4 maintaining this role throughout the experiment. This highlights that the swarm can settle into sub-optimal mode configurations in spatial settings but still achieve high productivity.
\end{itemize}

We demonstrate that our collective localization approach operates in complex physical settings involving sensor noise, slippage, dynamic interaction profiles, and unconstrained trajectories. We showcase an array of interesting emergent sub-behavior that depends on local interactions: isolated robots individually mode-switch; robots with many interactions designate perfect localizers; robots adapt to changing interaction rates. These emergent behaviors collectively boost the productivity of the swarm. As collective localization is compatible with unconstrained trajectories, it can be applied to numerous inspection scenarios. 

\section{Conclusion and Future Work}
Inspired by nature, we introduce a self-organized approach to collaborative localization that enhances inspection productivity of a robot swarm. This is achieved by strategically sacrificing some individuals to optimize computational efficiency. We showcase the advantages of this approach using theoretical models, numerical simulations, and hardware experiments conducted with a swarm of 10 metal-climbing robots. Our analytical mean-field model demonstrates that sacrificing agents as perfect localizers can improve the productivity of the swarm and that the swarm can dynamically reconfigure itself to maximize productivity for any interaction regime. This collaborative strategy depends only on local interactions and scales for any number of agents. We can further boost productivity using smart collaboration, as validated using numerical simulations. Our hardware experiments demonstrate the applicability and robustness of this collective localization mechanism in complex inspection-like scenarios. Future work involves running hardware experiments using more sophisticated metal climbing robots with vision and SLAM capabilities. We also plan to generalize this approach to scenarios beyond localization, such as vigilance and obstacle avoidance.

\bibliography{CamReady} 

\section{Acknowledgments}
This work is supported by NSF CMMI-2036359 and Amazon Robotics Research Award. Hungtang Ko is supported by the James S. McDonnell Foundation’s Postdoctoral Fellowship. We thank Dr. Merihan Alhafnawi, Darren Chiu, Thiemen Siemensma, and Prof. Bahar Haghighat for assistance with the hardware testbed. 

Inspired by a recent initiative to increase awareness and mitigate citation bias \cite{zurn2020citation}, we include a gender citation diversity statement using manually compiled data. Our references contain 12.1\% woman (first author)/woman (last author), 18.2\% man/woman, 6.1\% woman/man, and 63.6 \% man/man. The statistics we report fail to include intersex, non-binary, and transgender people.
We look forward to future work that could help us better understand how to support equitable practices in science.

\end{document}